\documentclass[journal,onecolumn]{IEEEtran}
\ifCLASSINFOpdf
\else
\fi
\usepackage{epsfig}
\usepackage{graphicx}
\usepackage{amsmath}
\usepackage{amssymb}
\usepackage{commath}
\usepackage{bbding}
\usepackage{cite}
\usepackage{cleveref}
\usepackage{booktabs,multirow}

\usepackage{times}

\usepackage{mathtools}
\DeclarePairedDelimiter\ceil{\lceil}{\rceil}
\DeclarePairedDelimiter\floor{\lfloor}{\rfloor}

\begin{document}
%
\title{Volumetric Super-Resolution of Multispectral Data}
%
%
%

\author{Vildan~Atalay~Aydin and~Hassan~Foroosh
\thanks{Vildan Atalay Aydin and Hassan Foroosh are with the Department of Computer Science, University of Central Florida, Orlando,
FL, 32816 USA (e-mails: vatalay@knights.ucf.edu and foroosh@cs.ucf.edu).}
}

\maketitle

\begin{abstract}
Most multispectral remote sensors (e.g. QuickBird, IKONOS, and Landsat 7 ETM+) provide low-spatial high-spectral resolution multispectral (MS) or high-spatial low-spectral resolution panchromatic (PAN) images, separately. In order to reconstruct a high-spatial/high-spectral resolution multispectral image volume, either the information in MS and PAN images are fused (i.e. pansharpening) or super-resolution reconstruction (SRR) is used with only MS images captured on different dates. Existing methods do not utilize temporal information of MS and high spatial resolution of PAN images together to improve the resolution. In this paper, we propose a multiframe SRR algorithm using pansharpened MS images, taking advantage of both temporal and spatial information available in multispectral imagery, in order to exceed spatial resolution of given PAN images. We first apply pansharpening to a set of multispectral images and their corresponding PAN images captured on different dates. Then, we use the pansharpened multispectral images as input to the proposed wavelet-based multiframe SRR method to yield full volumetric SRR. The proposed SRR method is obtained by deriving the subband relations between multitemporal MS volumes. We demonstrate the results on Landsat 7 ETM+ images comparing our method to conventional techniques.
\end{abstract}

\begin{IEEEkeywords}
Volumetric Multispectral Super-resolution, Discrete Wavelet Transforms \and Haar Domain Subband Relations
\end{IEEEkeywords}

\section{Introduction} \label{sec:intro}
Super-resolution and image restoration methods \cite{Foroosh_2000,Foroosh_Chellappa_1999,Foroosh_etal_1996,Cao_etal_2015,berthod1994reconstruction,shekarforoush19953d,lorette1997super,shekarforoush1998multi,shekarforoush1996super,shekarforoush1995sub,shekarforoush1999conditioning,shekarforoush1998adaptive,berthod1994refining,shekarforoush1998denoising,bhutta2006blind,jain2008super,shekarforoush2000noise,shekarforoush1999super,shekarforoush1998blind} have a wide-range of applications in various areas of imaging and computer vision, such as self-localization \cite{Junejo_etal_2010,Junejo_Foroosh_2010,Junejo_Foroosh_solar2008,Junejo_Foroosh_GPS2008,junejo2006calibrating,junejo2008gps}, image annotation \cite{Tariq_etal_2017_2,Tariq_etal_2017,tariq2013exploiting,tariq2015feature,tariq2014scene,tariq2015t}, surveillance \cite{Junejo_etal_2007,Junejo_Foroosh_2008,Sun_etal_2012,junejo2007trajectory,sun2011motion,Ashraf_etal2012,sun2014feature,Junejo_Foroosh2007-1,Junejo_Foroosh2007-2,Junejo_Foroosh2007-3,Junejo_Foroosh2006-1,Junejo_Foroosh2006-2,ashraf2012motion,ashraf2015motion,sun2014should}, action recognition \cite{Shen_Foroosh_2009,Ashraf_etal_2014,Ashraf_etal_2013,Sun_etal_2015,shen2008view,sun2011action,ashraf2014view,shen2008action,shen2008view-2,ashraf2013view,ashraf2010view,boyraz122014action,Shen_Foroosh_FR2008,Shen_Foroosh_pose2008,ashraf2012human}, object tracking \cite{Shu_etal_2016,Milikan_etal_2017,Millikan_etal2015,shekarforoush2000multi,millikan2015initialized}, shape description and object recognition \cite{Cakmakci_etal_2008,Cakmakci_etal_2008_2,Zhang_etal_2015,Lotfian_Foroosh_2017,Morley_Foroosh2017,Ali-Foroosh2016,Ali-Foroosh2015,Einsele_Foroosh_2015,ali2016character,Cakmakci_etal2008,damkjer2014mesh}, scene modeling \cite{Cakmakci_etal_2008,Cakmakci_etal_2008_2,Zhang_etal_2015,Lotfian_Foroosh_2017,Morley_Foroosh2017,Ali-Foroosh2016,Ali-Foroosh2015,Einsele_Foroosh_2015,ali2016character,Cakmakci_etal2008,damkjer2014mesh}, image-based rendering \cite{Cao_etal_2005,Cao_etal_2009,shen2006video,balci2006real,xiao20063d,moore2008learning,alnasser2006image,Alnasser_Foroosh_rend2006,fu2004expression,balci2006image,xiao2006new,cao2006synthesizing}, and camera motion estimation \cite{Cao_Foroosh_2007,Cao_Foroosh_2006,Cao_etal_2006,Junejo_etal_2011,cao2004camera,cao2004simple,caometrology,junejo2006dissecting,junejo2007robust,cao2006self,foroosh2005self,junejo2006robust,Junejo_Foroosh_calib2008,Junejo_Foroosh_PTZ2008,Junejo_Foroosh_SolCalib2008,Ashraf_Foroosh_2008,Junejo_Foroosh_Givens2008,Lu_Foroosh2006,Balci_Foroosh_metro2005,Cao_Foroosh_calib2004,Cao_Foroosh_calib2004,cao2006camera}
to name a few. The goal of multiview super-resolution is to reconstruct a high resolution (HR) image by fusing a sequence of degraded or aliased low resolution (LR) images of the same scene, where degradation can be a consequence of motion, camera optics, atmospheric distortions, undersampling, etc. 
\\

Multispectral sensors typically provide low-spatial high-spectral resolution for the multispectral (MS) volume, and high-spatial low-spectral resolution for the panchromatic (PAN) images. This is often due to technological limitations inherent in satellite sensors. However, numerous remote sensing applications related to land-cover management, environmental monitoring, weather forecasting, and topographic map updating require high-spatial high-spectral resolution MS image volume.

\begin{figure}
\begin{center}
\includegraphics[width=0.6\linewidth]{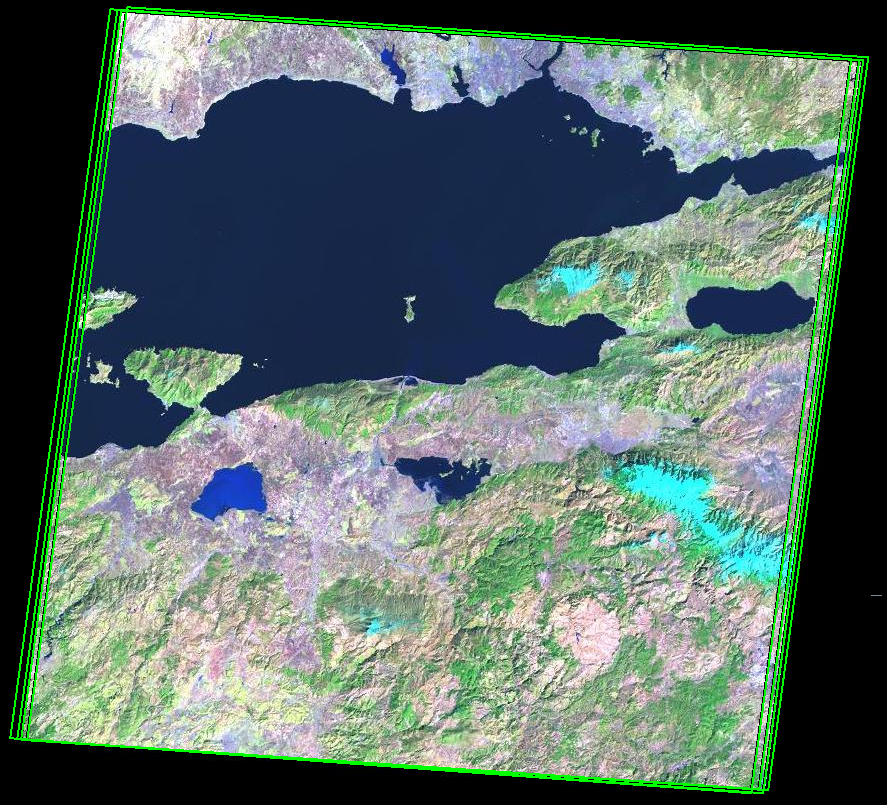}
\caption{Landsat 7 ETM+ multispectral images taken at different dates (shown in green areas) around Sea of Marmara.}
\end{center}
\end{figure} \label{fig:istanbul}

In order to obtain high spatial resolution MS images, a large body of research is devoted to fusing information of MS and PAN bands, which is called pansharpening. These methods do not consider the temporal information captured by the sensors. Multiframe SRR methods, on the other hand, fuse a sequence of degraded or aliased low resolution (LR) images of the same scene taken at different times, from different viewpoints or by different sensors, to obtain a high resolution (HR) image. Fig. \ref{fig:istanbul} shows an example of multispectral images taken at different times by the same sensor (i.e. Landsat 7 ETM+). Approaches to solve the SRR problem can be classified into frequency domain, interpolation, regularization, and learning-based methods \cite{tian2011survey}.

Fourier-based methods \cite{tsai,robinson2010efficient,vandewalle2007super} use the aliasing property of LR images in order to reconstruct an HR image. Although these methods are intuitive and have low computational complexity, due to their global nature, they only allow linear space invariant blur (PSF); and it is difficult to identify a global frequency-domain {\em a priori} knowledge to overcome ill-posedness. Some examples of Fourier domain techniques include the works by; Tsai and Huang \cite{tsai} which exploits the relationship between Continuous Fourier Transform (CFT) of the unknown HR scene and Discrete Fourier Transform (DFT) of the shifted and sampled LR images, Robinson et al.~ \cite{robinson2010efficient} which applies combined Fourier-wavelet de-convolution and de-noising algorithm, and Vandewalle et al.~ \cite{vandewalle2007super} where joint registration and reconstruction is performed using multiple unregistered images.
Spatial-domain interpolation-based methods \cite{irani1991improving,zhou2012interpolation} tackle Fourier-domain obstacles by fusing the information from all LR images using a general interpolation technique (e.g. nearest neighbor, bilinear, bicubic). However, these methods result in overly smoothed images. As examples of interpolation based methods, Irani and Peleg \cite{irani1991improving} update the HR estimate by iteratively back-projecting the difference between the approximation and exact image, and Zhou et al.~ \cite{zhou2012interpolation} utilize multi-surface fitting. 
In order to stabilize the ill-posed problem of SRR, regularization-based methods \cite{marquina2008image,farsiu2006multiframe} optimize a cost function with a regularization term by incorporating prior knowledge, where probabilistic estimators such as the maximum likelihood \cite{elad2001fast}, maximum {\em a posteriori} (MAP) \cite{shen2007map}, and Bayesian \cite{babacan2011variational,bishop2006bayesian} can be employed. The drawbacks and challenges in these methods are determination of the prior model, high computational cost, and over-smoothing. More examples of regularization-based methods include the works by Marquina and Osher \cite{marquina2008image} that employs Bregman iteration for Total Variation Regularization, and Farsiu et al.~ \cite{farsiu2006multiframe} in which a multi-term cost function is minimized. 
Finally, learning based methods \cite{freeman2002example,zhang2011partially,wang2015self,glasner2009super} obtain an HR image from a single image by utilizing training sets of LR/HR images or patch pairs. The problems with these methods include their high computational cost, and correspondence ambiguities between HR and LR images. Clustering and supervised neighbor embedding is employed by Zhang et al.~ \cite{zhang2011partially}. Moreover, Glasner et al.~ \cite{glasner2009super} combine multi-image super resolution and example-based approaches based on the assumption that patches in natural images recur many times inside the image.
In order to overcome the drawbacks of aforementioned methods, recent research in SRR explores wavelet-based techniques \cite{ji2009robust, robinson2010efficient, demirel2011discrete, dong2011image}). The intuition behind these approaches is that the LR images can be used to model the lowpass subband of the unknown HR image, in order to reconstruct the high frequency information lost during image acquisition. 

Li et al.~ \cite{li2009superresolution} propose a wavelet-based multiframe SRR technique, where temporal information of low-spatial-resolution MS images are used to increase their spatial resolution, as opposed to the pansharpening methods, where PAN images are employed instead of temporal data. They propose using an SRR method based on MAP with a universal Hidden Markov Tree model as a preprocessing step  for multispectral image classification, where SRR is applied band-by-band to multispectral images captured on different dates. However, although their method uses the temporal information, it does not take advantage of the high spatial resolution PAN image available with most multispectral sensors.

In this paper, we propose a wavelet-based multiframe SRR method that takes advantage of both temporal and spatial information captured by multispectral sensors, in order to obtain a higher spatial resolution MS image, which exceeds the spatial resolution of the available PAN image, while preserving the high spectral resolution. To this end, we first apply pansharpening to a set of multispectral and corresponding PAN images taken at different times, in order to obtain high spatial resolution MS images. We then use our proposed wavelet-based approach for SRR band-by-band on the pansharpened MS images, in order to achieve a higher spatial resolution MS image. 
In addition to simultaneous pansharpening and SRR our work makes the following contributions: (i) We assume that pansharpened MS images (i.e. LR) correspond to the approximation coefficients of the first level discrete wavelet transform (DWT) of an unknown higher spatial resolution MS image (i.e. HR) and use their second level detail coefficients for initialization (initial estimate of the high frequency content of the HR image); (ii) We establish explicit closed-form expressions that define how the local high-frequency information that we aim to recover for the HR image are related to the local low frequency information in the given sequence of LR views (i.e. wavelet domain in-band relations across LR images). The derived formulae are provided utilizing the Haar DWT due to their simplicity and low computational requirements.  However, a general formulation for other types of wavelets can be readily derived in a similar manner. (iii) We solve the newly formulated SRR problem in an iterated back-projection optimization framework that focuses solely on recovering the high-frequency information across multiple pansharpened images, i.e. recovering only the wavelet detail coefficients of the HR image (unlike typical multi-frame methods that try to recover the HR image at all frequencies by fusion of LR images). In that sense, our method is similar to single-image SRR, while still extracting information across multiple images. Our superior results can be attributed to the exactness (closed-form), well-posedness, and the linearity of the equations derived in Section \ref{sec:shift}, and the inherent nature of wavelets, making them very effective in simultaneous signal information localization in space-time-frequency.

The remainder of this paper is organized as follows. In Section \ref{sec:related}, a brief summary of related research is provided. The pansharpening method employed in this paper is explained in Section \ref{sec:pansharpening}. Sections \ref{sec:shift} and \ref{sec:sr} present the derived closed-form linear relationships for wavelet coefficients, and the proposed approach for SRR, respectively. In Section \ref{sec:exp}, we present the experimental results and the comparisons. Finally, we conclude our paper in Section \ref{sec:conc} with some closing remarks.

\section{Related Work} \label{sec:related}
 
Wavelet-based SRR approaches can be summarized as follows. In order to reduce noise in SRR methods, Robinson et al.~ \cite{robinson2010efficient} apply a combined Fourier-wavelet deconvolution and denoising algorithm to multiframe SRR. On the other hand, to reduce degradation artifacts such as blurring and the ringing effect, Temizel and Vlachos \cite{temizel2005wavelet} utilize zero padding in the wavelet domain followed by cycle spinning. Zhao et al.~ \cite{zhao2003wavelet} solve a constrained optimization problem utilizing wavelet domain Hidden Markov Tree (HMT) model to solve the prior knowledge problem, since HMT characterizes the statistics of real world images accurately. 
Nguyen and Milanfar \cite{nguyen2000efficient}, unlike the conventional interpolation based methods, use the regularity and structure in the interlaced sampling of LR images. For deblurring,  Chan et al.~ \cite{chan2003wavelet} derive iterative algorithms, which decompose the HR image obtained from an iteration into different frequency components and add them to the next iteration. Ji and Fermuller \cite{ji2009robust} handle image registration and reconstruction together, by first estimating the homographies between multiple images, then reconstruct the HR image in a wavelet-based iterative back-projection scheme. Jiji et al.~ \cite{jiji2004single}, as an example to learning-based methods, handle the problem of representing the relationship between LR-HR frames with training their dataset with HR images by learning from wavelet coefficients at finer scales, followed by regularization in a least-squares manner. The work in \cite{gajjar2010new} follows Jiji's method and employs discrete wavelet transform for training, where a cost function based on MAP estimation is optimized with gradient descent method, employing an Inhomogeneous Gaussian Markov random field prior. Dong et al.~ \cite{dong2011image} learn various sets of bases from a precollected dataset of example image patches, and select one set of bases adaptively to characterize the local sparse wavelet domain. 
The above stated methods are performed either iteratively which requires high computational time or based on interpolation which results in overly smooth images. Our goal is to derive a direct relationship between LR images for a closed-form SRR solution, which prevents sacrificing high quality. 

SRR methods are widely utilized for remote sensing. Demirel and Anbarjafari \cite{demirel2011discrete} use DWT and an intermediate stage for estimating high frequency information for satellite image super-resolution. Patel and Joshi \cite{patel2015super} propose a learning-based approach for SRR of hyperspectral images using the DWT, where application-specific wavelet basis (i.e. filter coefficients) are designed. Moreover, Zhang et al.~ \cite{zhang2012super} propose a MAP-based multiframe SRR method for hyperspectral images, where PCA is employed in order to reduce the computational complexity.

\section{Pansharpening Technique} \label{sec:pansharpening}
Based on the main technique used, pansharpening methods can be categorized into five groups \cite{amro2011survey} as component substitution, relative spectral contribution, high-frequency injection, image statistics based, and multiresolution. Component substitution methods upsample and transform MS images and substitute components of the transformed MS bands with components from the PAN images (e.g. Intensity-Hue-Saturation (IHS) pansharpening
, principal component substitution
, and Gram-Schmidt spectral sharpening
). Relative spectral contribution approach, on the other hand, employ a linear combination of bands instead of using substitution (e.g. Brovey transform
). High-frequency injection techniques transfer the high frequency content of the PAN image to the MS images (e.g. the High Pass Modulation method
); whereas image statistics based models use the statistical relationship between each band of the MS and PAN images (e.g. Bayesian based techniques 
and Price's method
). Finally, multiresolution methods decompose MS and PAN images into different spatial levels to demonstrate the relationship between PAN and MS images in coarser scales and improve spatial details (e.g. Laplacian pyramid
, wavelet
, and contourlet based 
methods). 

Wavelet-based pansharpening methods include the works by Zhou et al.~ \cite{zhou1998wavelet} which merge images by performing an inverse DWT using the approximation image from each Landsat TM band and detail images from SPOT PAN, the work by Kim et al.~ \cite{kim2011improved}, where an improved additive-wavelet fusion method is proposed using the \`{a} trous algorithm which does not decompose the MS image and inject high frequency following the frequency of the MS image using a low-resolution PAN image; and the work by Ranchin et al.~ \cite{ranchin2003image} where two multiscale models and two inter-band structure models are described for ARSIS concept. Alparone et al.~ \cite{alparone2007comparison} compare several methods and conclude that the multiresolution based ones and the methods that employ adaptive models for the injection of highpass details outperform all the others.

Based on the comparison results obtained by Alparone et al.~ \cite{alparone2007comparison} and Bovolo et al.~ \cite{bovolo2010analysis}, and the nature of the proposed method in terms of wavelet domain relations described in the next section, the Additive Wavelet Luminance Proportional (AWLP) method \cite{otazu2005introduction} is chosen for our framework. The AWLP method is an extended version of the additive wavelet luminance technique (AWL), which is designed for three-band (RGB) multispectral images and works in the IHS domain. The AWL method injects high frequency information of the PAN image to MS images proportional to their original values in order to preserve the radiometric signature of MS images. The AWLP method generalizes the AWL method to include arbitrary number of bands, as follows:

\begin{equation}
HR_i = LR_i + \dfrac{LR_i}{\sum_{i=1}^{L}LR_i} \sum_{j=1}^{n}w_{PAN}
\end{equation}

\noindent where LR and HR are low and high spatial resolution MS images, respectively, $L$ is the number of bands, $n$ is the number of DWT decomposition levels, and $w_{PAN}$ is the DWT decomposition of the PAN image.

\section{Inter-subband Relationships} \label{sec:shift}
In this section, the relationships between wavelet subbands of multitemporal images are derived in terms of in-band (wavelet domain) shifts of the reference image subbands. We first summarize the notation used throughout the paper.

\subsection{Notation} \label{term}

Here, we provide the notation used, in Table \ref{termtable}. 
	\begin{table}[h] 
		\centering
		\small
		\caption{Notation}
		\begin{tabular}{l p{0.7\linewidth}}
			$I$ & Reference HR image\\
			${\bf A}, {\bf a}, {\bf b}, {\bf c}$ & Haar wavelet transform approximation, horizontal, vertical, and diagonal subbands of HR image, respectively \\
			${\bf F}, {\bf K}, {\bf L}$ & Coefficient matrices to be multiplied by approximation, horizontal, vertical, and diagonal DWT subbands \\
			$h$ & Number of hypothetically added levels in case of non-integer shifts\\
			$s$ & Integer shift amount after the hypothetically added levels ($h$)\\
		\end{tabular} \label{termtable}
	\end{table} 

Bold letters in the following sections demonstrate matrices and vectors. Subscripts $x$ and $y$ indicate the horizontal and vertical translation directions, respectively. 

\subsection{In-band Shifts}
As a wavelet-based SRR method, our goal is to reconstruct an HR image (i.e. HR MS) using the given LR sequence (i.e. pansharpened MS images) assumed to represent the lowpass subbands of an unknown HR sequence of images. Therefore, we derive the relationships between the DWT subbands of the reference images and the target HR images, assuming that their lowpass subbands are the given LR images for SRR. To derive the closed-form relationships, we assume that HR images are given, and describe in-band shifts (in the wavelet domain) of the reference HR image subbands. For the SRR method, we inverse the process and use these relationships as our model, in order to estimate the high-frequency information of the unknown reference HR image, given lowpass subbands. Below, we derive the mathematical expressions which demonstrate these relationships.

Let $A$, $a$, $b$, and $c$ be the first level approximation, horizontal, vertical, and diagonal detail coefficients (subbands), respectively, of a $2D$ reference HR image, $I$, of size $2m\times2n$, where $m$ and $n$ are positive integers. Since decimation operator in DWT reduces the size of input image by half in each direction for each subband, we require the image sizes to be divisible by 2. Now, the $1st$ level Haar transform subbands of the target HR image translated in any direction (i.e. horizontal, vertical, diagonal), can be expressed in matrix form using the $1st$ level subbands of the reference HR image as in the following equations (also illustrated in Fig. \ref{fig:shift}):

\begin{figure}[t]
	\centering
	\centerline{\includegraphics[width=10cm]{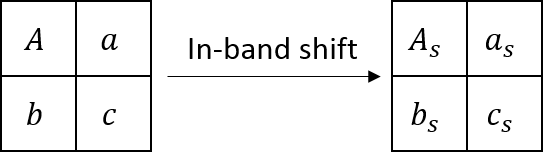}}
	\caption{In-band shift of reference image subbands.}\medskip \label{fig:shift}
\end{figure}

\begin{eqnarray}\label{firsteq} 
{\bf A}_s &=& {\bf F}_y {\bf A} {\bf F}_x + {\bf F}_y {\bf a} {\bf K}_1 + {\bf L}_1 {\bf b} {\bf F}_x + {\bf L}_1 {\bf c} {\bf K}_1 \nonumber\\ 
{\bf a}_s &=& - {\bf F}_y {\bf A} {\bf K}_1 + {\bf F}_y {\bf a} {\bf K}_2 - {\bf L}_1 {\bf b} {\bf K}_1 + {\bf L}_1 {\bf c} {\bf K}_2 \nonumber\\
{\bf b}_s &=& - {\bf L}_1 {\bf A} {\bf F}_x - {\bf L}_1 {\bf a} {\bf K}_1 + {\bf L}_2 {\bf b} {\bf F}_x + {\bf L}_2 {\bf c} {\bf K}_1 \nonumber\\
{\bf c}_s &=& {\bf L}_1 {\bf A} {\bf K}_1 - {\bf L}_1 {\bf a} {\bf K}_2 - {\bf L}_2 {\bf b} {\bf K}_1 + {\bf L}_2 {\bf c} {\bf K}_2 \nonumber\\
\end{eqnarray} 

As already mentioned in Section \ref{term}, ${\bf F}$, ${\bf K}$, and ${\bf L}$ stand for coefficient matrices to be multiplied by the lowpass and highpass subbands of the reference HR image, where subscripts $x$ and $y$ indicate \textit{horizontal} and \textit{vertical} shifts. ${\bf A}_s, {\bf a}_s, {\bf b}_s, {\bf c}_s$ are translated (i.e. target) HR image subbands in any direction. The low/highpass subbands of both reference and target images are of size $m \times n$, ${\bf F}_y$ and ${\bf L}_{1,2}$ are $m \times m$, whereas ${\bf F}_x$ and ${\bf K}_{1,2}$ are $n \times n$.

By examining the translational shifts between subbands of two input frames in the Haar domain, we realize that horizontal translation reduces ${\bf L}$ to zero and ${\bf F}_y$ to the identity matrix. This could be understood by examining the coefficient matrices defined later in this section (namely, Eq. (\ref{coefmat})), by setting the related vertical components to zero (specifically, $s_y$ and $h_y$). Likewise, vertical translation depends solely on approximation and vertical detail coefficients, in which case ${\bf K}$ is reduced to zero and ${\bf F}_x$ is equal to the identity matrix. 

Here, we first define the matrices for subpixel shift amounts. The algorithm to reach any shift amount using the subpixel relationship will be described later in this section.

\begin{figure}[ht]
	\centering
	\centerline{\includegraphics[width=12cm]{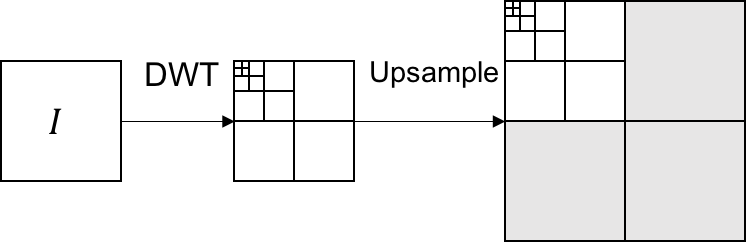}}
	\caption{Illustration of upsampling process.}\medskip \label{fig:upsample}
\end{figure}

Contrary to the customary model of approximating a subpixel shift by upsampling an image followed by an integer shift, our method models subpixel shift directly based on the original coefficients of the reference HR image, without upsampling and the need for interpolation. To this end, we resort to the following observations: 

{\bf (1)} Upsampling an image $I$, is equivalent to adding levels to the bottom of its wavelet transform, and setting the detail coefficients to zero, while the approximation coefficients remain the same, as demonstrated in Fig. \ref{fig:upsample} for upsampling by $2^1$ as an example, where gray subbands show added zeros. 

{\bf (2)} Shifting the upsampled image by an amount of $s$ is equivalent to shifting the original image by an amount of $s/2^h$, where $h$ is the number of added levels (e.g. $h=1$ in Fig. \ref{fig:upsample}). 

These observations allow us to perform in-band shift of the reference image subbands for a subpixel amount, without upsampling or interpolation. 
Transformed images, therefore, can be found by using the original level subbands of the reference HR image with the help of a hypothetically added level ($h$) and an integer shift value ($s$) at the added level.

Now, the aforementioned coefficient matrices, ${\bf F}_x$, ${\bf K}_{1}$, and ${\bf K}_{2}$ can be defined, in lower bidiagonal Toeplitz matrix form as follows.

\begin{eqnarray} 
	{\bf F}_x = \dfrac{1}{2^{h_x+1}} 
	\begin{bmatrix}
	2^{h_x+1} - \abs{s_x} & &   \\
	\abs{s_x} & 2^{h_x+1} - \abs{s_x}   \\
	& \abs{s_x} \\
	& & \ddots & \ddots \\
	& \\
	& & & \abs{s_x} & 2^{h_x+1} - \abs{s_x} \\ \nonumber
	\end{bmatrix} \nonumber
\end{eqnarray} 

\begin{eqnarray} 
{\bf F}_y = \dfrac{1}{2^{h_y+1}} 
	\begin{bmatrix}
	2^{h_y+1} - \abs{s_y} & \abs{s_y}   \\
	& 2^{h_y+1} - \abs{s_y} & \abs{s_y}   \\
	& & \ddots & \ddots\\
	& \\
	& & & & \abs{s_y} \\
	\end{bmatrix}
\end{eqnarray} 

\begin{eqnarray}
{\bf K}_1 = \dfrac{1}{2^{h_x+1}} 
\begin{bmatrix}
-s_x &  \\
s_x & -s_x   \\
& s_x & \\
& & \ddots & \ddots\\
&\\
& & & s_x & -s_x \\ 
\end{bmatrix} \nonumber
\end{eqnarray}

\begin{eqnarray}
{\bf L}_1 = \dfrac{1}{2^{h_y+1}} 
\begin{bmatrix}
-s_y & s_y  & & & \\
& -s_y & s_y   \\
& & \ddots & \ddots\\
& & &  -s_y &  s_y \\
& & & & -s_y \\ 
\end{bmatrix} \nonumber
\end{eqnarray} 

\begin{eqnarray}
{\bf K}_2 = \dfrac{1}{2^{h_x+1}} 
	\begin{bmatrix}
	2^{h_x+1} - 3\abs{s_x} &  \\
	- \abs{s_x} & 2^{h_x+1} - 3\abs{s_x}   \\
	& -\abs{s_x}   & \\
	& & \ddots & \ddots\\
	&\\
	& & & -\abs{s_x} & 2^{h_x+1} - 3\abs{s_x}  \\ 
	\end{bmatrix} \label{coefmat}
\end{eqnarray}

\begin{eqnarray}
{\bf L}_2 = \dfrac{1}{2^{h_y+1}} 
\begin{bmatrix}
2^{h_y+1} - 3\abs{s_y}  & -\abs{s_y} & & & \\
& 2^{h_y+1} - 3\abs{s_y} & -\abs{s_y}  \\
& & & \ddots & \ddots\\
& & & &  -\abs{s_y} \\
& & & & 2^{h_y+1} - 3\abs{s_y}  \\ 
\end{bmatrix} 
\end{eqnarray} 

\normalsize
\noindent where $s_{x}$ and $h_{x}$ demonstrate the integer shift amount at the hypothetically added level and the number of added levels for $x$ direction, respectively. 

${\bf F}_y$, ${\bf L}_1$, and ${\bf L}_2$ matrices are defined in a similar manner by upper bidiagonal Toeplitz matrices, using $y$ direction values for $s$ and $h$. When the shift amount is negative, diagonals of the coefficient matrices interchange.

As mentioned earlier, ${\bf F}_x$ and ${\bf K}_{1,2}$ are $n \times n$, while ${\bf F}_y$ and ${\bf L}_{1,2}$ are $m \times m$. Sizes of these matrices also indicate that in-band shift of subbands is performed using only the original level Haar coefficients (which are of size $m \times n$) without upsampling.  

The relationship defined above for subpixel shifts, can be used to produce any shift amount based on the fact that wavelet subbands are periodically shift-invariant. Table \ref{shifts} demonstrates the calculation of any shift using subpixels, where $\%$ stands for modulo, $\floor{s}$ and $\ceil{s}$ are the greatest integer lower than, and smallest integer higher than the shift, $s$. Performing circular shift of subbands for given values in each shift amount case, and setting the new shift amount to the new shifts in Table \ref{shifts}, we can calculate any fractional or integer amount of shifts using subpixels.

	\begin{table}[h] 
		\scriptsize
		\centering
		\caption{Arbitrary shifts based on circular shift and subpixel values}
		\begin{tabular}{lll}
			\toprule
			Shift amount & Circular shift & New shift amount\\
			\midrule
			$s\%2 = 0$ & $s/2$ 	& $0$\\	
			$s\%2 = 1$ & $\floor{s/2}$ & $1$ \\
			$\ceil{s}\%2 = 0$ & $\ceil{s}/2$ & $s-\ceil{s}$ \\
			$\floor{s}\%2 = 0$ & $\floor{s}/2$ & $s-\floor{s}$ \\
			\bottomrule	
		\end{tabular} \label{shifts}
	\end{table} 

If the shift amount (or the new shift in Table \ref{shifts}) is not divisible by $2$, in order to reach an integer value at the $(N+h)$th level, the shift value at the original level is rounded to the closest decimal point which is divisible by $2^h$.

\section{Super Resolution Reconstruction} \label{sec:sr}

In this section, we first present the SRR observation model, followed by our proposed method.

\subsection{Observation Model} \label{obs}

Let $I(\sigma m\times \sigma n)$ denote the desired HR image, and $A_k$ be the $k$th observed LR image, with a downsampling ration of $\sigma$. The SRR observation model is given by:

\begin{eqnarray}
{\bf A}_k = {\bf D}_k {\bf B}_k {\bf M}_k {\bf I} + {\bf n}_k,\quad k = 1,2,...,K
\end{eqnarray}

\noindent where ${\bf M}_k$, ${\bf B}_k$, ${\bf D}_k$, and ${\bf n}_k$ denote motion, blurring effect, downsampling operator, and noise term for the $k$th LR image, respectively. In the above formula, LR and HR images are rearranged in lexicographic order; therefore, ${\bf I}$ is of size $\sigma^2 mn\times 1$, ${\bf A}_k$ and ${\bf n}_k$ are $mn \times 1$, ${\bf B}_k$ and ${\bf M}_k$ are $\sigma^2 mn \times \sigma^2 mn$, and ${\bf D}_k$ is $mn \times \sigma^2 mn$.

Assuming the same downsampling ratio for all LR images, and taking bands of MS images into account, we can modify the above observation model as follows:

\begin{eqnarray}
{\bf A}_{k,i} = {\bf D} {\bf B}_{k,i} {\bf M}_{k,i} {\bf I} + {\bf n}_{k,i}
\end{eqnarray}

\noindent where $k = 1,2,...,K$, $i = 1,2,...,L$, $K$ is the number of LR images, and $L$ is the number of MS bands.

Given a sequence of pansharpened MS bands, ${\bf A}_{k,i}$, our goal is to reconstruct the unknown HR MS image, ${\bf I}$, which exceeds the spatial resolution of the available PAN image.

\subsection{Proposed Method}

\begin{figure}
\centering 
\centerline{\includegraphics[width=0.8\linewidth]{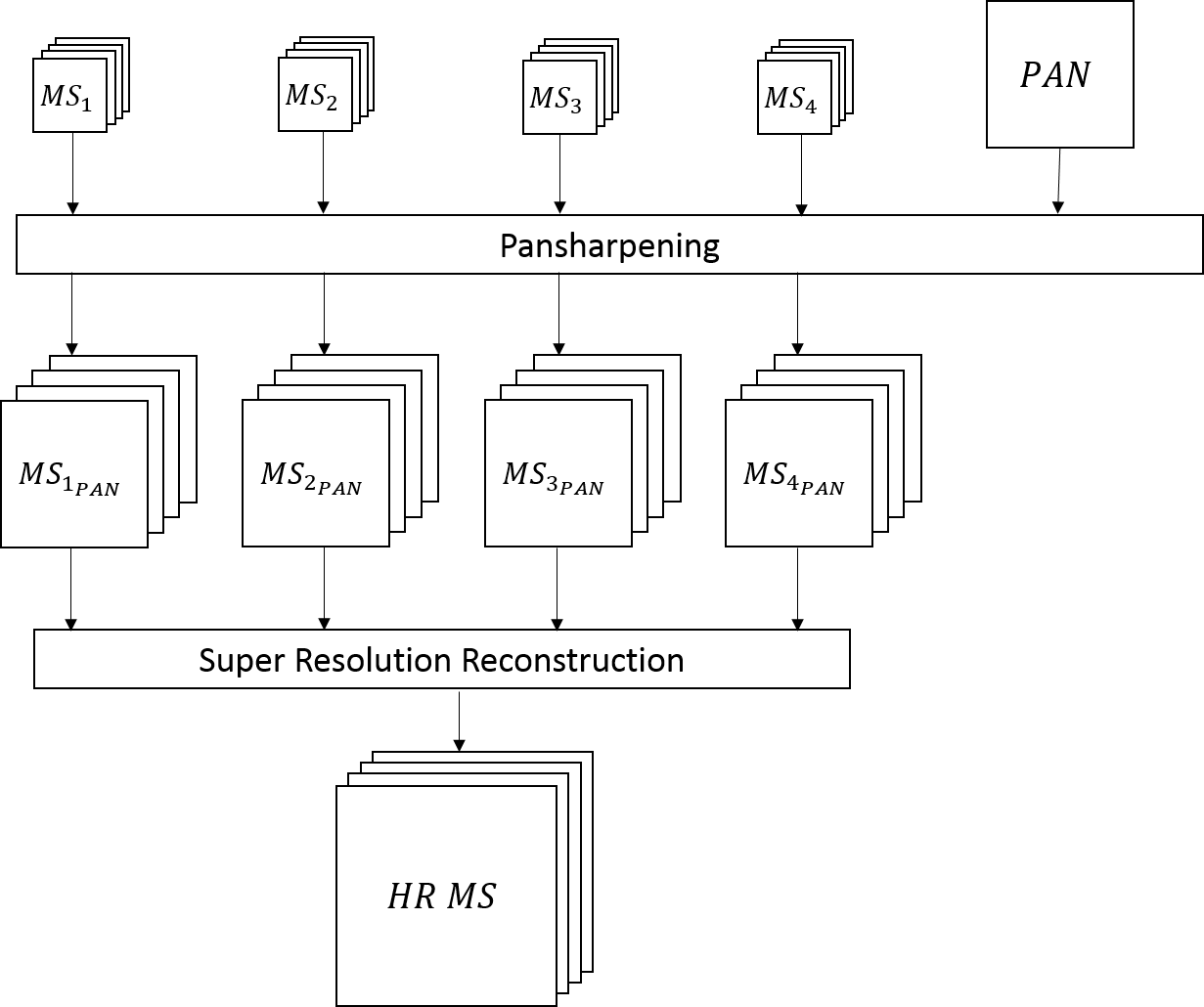}}
\caption{Flowchart of the proposed method.} \label{fig:flowchart}
\end{figure}

Spatial resolution of PAN images differ based on the multispectral sensors. For example, while Landsat 7 ETM+ sensors provide $15$ m spatial resolution PAN images, Quickbird sensors increase this amount upto $61$ cm. In order to increase the available spatial resolution of PAN images while keeping the spectral resolution provided by MS bands, we propose using both temporal and spectral information accessible via most multispectral sensors. Therefore, we first apply pansharpening to multispectral images captured around the same region at different dates. Then, using these pansharpened MS images, we perform multiframe SRR in order to exceed the spatial resolution of handy PAN images. Fig. \ref{fig:flowchart} shows a pictorial explanation of the flowchart of proposed scheme.

\begin{figure}
\centering 
\centerline{\includegraphics[width=\linewidth]{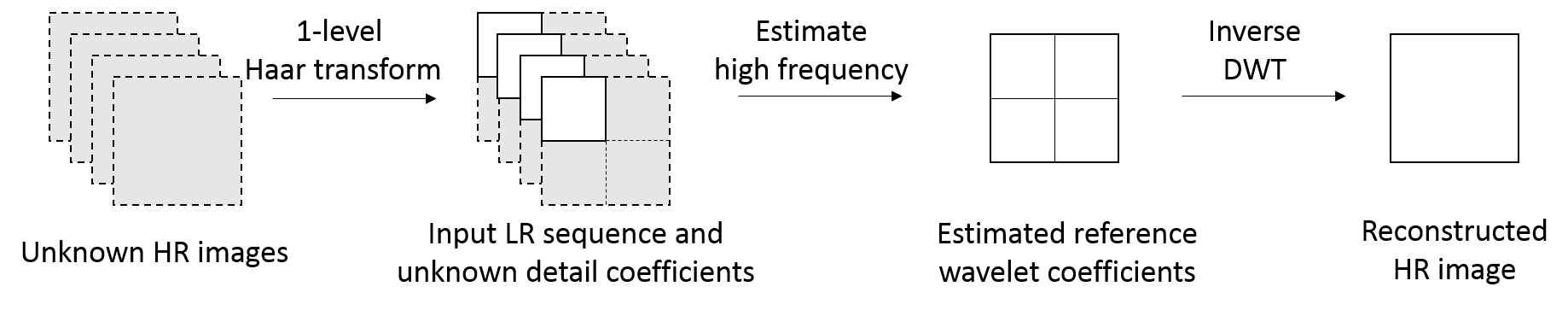}}
\caption{SRR for one band of MS image.} \label{fig:srr}
\end{figure} 

As in the underlying idea of wavelet-based SRR algorithms, we assume that the given LR image sequence (i.e. pansharpened MS images) is the lowpass subbands of unknown HR images. The goal is to reconstruct the unknown highpass subbands of one of these HR images, chosen as the reference. The SRR method described below is the inverse process of the method described in Section \ref{sec:shift}, where HR images are unknown, and high frequency information for one of these underlying HR images is estimated by a modified iterated back projection method. We perform the proposed scheme on each band of multitemporal pansharpened MS images. Fig. \ref{fig:srr} shows the SRR step for one band of MS images, where gray areas in dashed lines indicate unknown images and coefficients, and input LR sequence in white solid lines stand for one band of the pansharpened MS images taken at different times.

SRR process consists of two steps of image registration \cite{Foroosh_etal_2002,Foroosh_2005,Balci_Foroosh_2006,Balci_Foroosh_2006_2,Alnasser_Foroosh_2008,Atalay_Foroosh_2017,Atalay_Foroosh_2017-2,shekarforoush1996subpixel,foroosh2004sub,shekarforoush1995subpixel,balci2005inferring,balci2005estimating,foroosh2003motion,Balci_Foroosh_phase2005,Foroosh_Balci_2004,foroosh2001closed,shekarforoush2000multifractal,balci2006subpixel,balci2006alignment,foroosh2004adaptive,foroosh2003adaptive}
 and image reconstruction \cite{Foroosh_2000,Foroosh_Chellappa_1999,Foroosh_etal_1996,Cao_etal_2015,berthod1994reconstruction,shekarforoush19953d,lorette1997super,shekarforoush1998multi,shekarforoush1996super,shekarforoush1995sub,shekarforoush1999conditioning,shekarforoush1998adaptive,berthod1994refining,shekarforoush1998denoising,bhutta2006blind,jain2008super,shekarforoush2000noise,shekarforoush1999super,shekarforoush1998blind}. For the image registration step, we employed a method which uses SURF \cite{bay2006surf} and RANSAC \cite{fischler1981random}, implemented in IAT toolbox \cite{IAT2013}, where reference and target images are registered for affine transform. Since we assume that bands of an MS image captured by the same sensor are already aligned, we perform registration on one band only, and apply the same registration parameters to all bands. Once registration parameters are estimated, target images are rotated back for the estimated rotation amount in order for LR images to have only translational transform between them, since the formulae in Section \ref{sec:shift} are derived for translational shifts.

As a modification to the general Iterated Back Projection algorithm described first by Irani and Peleg \cite{irani1991improving}, we use the proposed method in Section \ref{sec:shift} to simulate motion. As in \cite{robinson2010efficient}, we change the order of motion and blur matrices in the SRR observation model as in:

\begin{eqnarray}
{\bf A}_{k,i} = {\bf D} {\bf M}_{k,i} {\bf B}_{k,i} {\bf I} + {\bf n}_{k,i}
\end{eqnarray}

Since our model uses wavelet decomposition of a reference image, in order to estimate the target images, the proposed method fuses the matrices ${\bf D}$ and ${\bf M}_{k,i}$: 

\begin{eqnarray}
{\bf A}_{k,i} = {\bf m}_{k,i} {\bf B}_{k,i} {\bf I} + {\bf n}_{k,i}
\end{eqnarray}

\noindent where ${\bf m}_{k,i} = {\bf D} * {\bf M}_{k,i}$ with $k = 1,2,...,K$, $i = 1,2,...,L$, as defined before.

High resolution image is then estimated by the following formula which is shown for one band:

\begin{equation}
\hat{{\bf I}}^{(n+1)} = \hat{{\bf I}}^{(n)} + \lambda \sum_{k} ({\bf A}_k - \hat{{\bf A}}_k) \textit{h}^{BP}  \label{eq:ibp}
\end{equation}

\noindent where $n$ stands for the iteration number, $\lambda$ is a step size, $\hat{{\bf A}}_k$ is estimated using current $I$ and motion parameters, and $\textit{h}^{BP}$ is a back projection kernel as defined in \cite{irani1991improving}.

The relationship between two translated LR images (i.e. ${\bf A}$ and ${\bf A}_s$ in Section \ref{sec:shift}) depends on the highpass subband of the reference HR image (i.e. ${\bf a}, {\bf b}, {\bf c}$). This fact is used to construct a linear system of equations based on known LR images and unknown highpass subbands of the reference HR image using related formulae from Eq. (\ref{firsteq}) for each target LR image. Since there are three unknowns (horizontal, vertical, and diagonal detail coefficients of the unknown HR image), at least three shifted LR images together with the reference LR image are required to solve the linear system. In order to avoid instabilities caused by inversion, the system is solved in an iterative least squares fashion. The linear system can be constructed as follows:

\begin{eqnarray}\label{eq:linear} 
{\bf A}_{s_1} &=& {\bf F}_{y_1} {\bf A} {\bf F}_{x_1} + {\bf F}_{y_1} {\bf a} {\bf K}_{1_1} + {\bf L}_{1_1} {\bf b} {\bf F}_{x_1} + {\bf L}_{1_1} {\bf c} {\bf K}_{1_1} \nonumber\\ 
{\bf A}_{s_2} &=& {\bf F}_{y_2} {\bf A} {\bf F}_{x_2} + {\bf F}_{y_2} {\bf a} {\bf K}_{1_2} + {\bf L}_{1_2} {\bf b} {\bf F}_{x_2} + {\bf L}_{1_2} {\bf c} {\bf K}_{1_2} \nonumber\\ 
{\bf A}_{s_3} &=& {\bf F}_{y_3} {\bf A} {\bf F}_{x_3} + {\bf F}_{y_3} {\bf a} {\bf K}_{1_3} + {\bf L}_{1_3} {\bf b} {\bf F}_{x_3} + {\bf L}_{1_3} {\bf c} {\bf K}_{1_3} \nonumber\\ 
\end{eqnarray} 

By rearranging (\ref{eq:linear}), we find:

\begin{equation} \label{eq:kron}
{\bf A}_{s_i} = ({\bf F}_{y_i} \otimes {\bf F}_{x_i}') {\bf A} + ({\bf F}_{y_i} \otimes {\bf K}_{1_i}') {\bf a} + ({\bf L}_{1_i} \otimes {\bf F}_{x_i}') {\bf b} + ({\bf L}_{1_i} \otimes {\bf K}_{1_i}') {\bf c} 
\end{equation} 

\noindent where $i = 1,2,3$. Here, ${\bf A}_{s_i}, {\bf A}, {\bf a}, {\bf b}$, and ${\bf c}$ are rearranged as $mn\times 1$ vectors, Kronecker tensor products in parenthesis are $mn \times mn$. 

In order to solve the system in Eq. (12) in a least squares manner, by reorganizing, we have;

\begin{equation}\label{eq:leastsquares} 
\begin{bmatrix}
{\bf A}_{s_1}\\
{\bf A}_{s_2}\\
{\bf A}_{s_3}\\
\end{bmatrix}
 =
\begin{bmatrix}
{\bf F}_{y_1} \otimes {\bf F}_{x_1}'\\
{\bf F}_{y_2} \otimes {\bf F}_{x_2}'\\
{\bf F}_{y_3} \otimes {\bf F}_{x_3}'\\
\end{bmatrix} {\bf A} + 
\begin{bmatrix}
{\bf F}_{y_1} \otimes {\bf K}_{1_1}' \quad {\bf L}_{1_1} \otimes {\bf F}_{x_1}' \quad  {\bf L}_{1_1} \otimes {\bf K}_{1_1}'\\
{\bf F}_{y_2} \otimes {\bf K}_{1_2}' \quad {\bf L}_{1_2} \otimes {\bf F}_{x_2}' \quad {\bf L}_{1_2} \otimes {\bf K}_{1_2}'\\
{\bf F}_{y_3} \otimes {\bf K}_{1_3}' \quad {\bf L}_{1_3} \otimes {\bf F}_{x_3}' \quad {\bf L}_{1_3} \otimes {\bf K}_{1_3}'\\
\end{bmatrix} 
\begin{bmatrix}
{\bf a} \\
{\bf b} \\
{\bf c}\\
\end{bmatrix} 
\end{equation} 

Finally, the system in Eq. (13) is solved by minimizing the following cost function:

\begin{equation} \label{eq:minimize}
\begin{bmatrix}
\hat{{\bf a}} \\
\hat{{\bf b}} \\
\hat{{\bf c}} \\
\end{bmatrix} 
\\
= \arg \min
\norm{
\begin{bmatrix}
{\bf F}_{y_1} \otimes {\bf K}_{1_1}' \quad {\bf L}_{1_1} \otimes {\bf F}_{x_1}' \quad  {\bf L}_{1_1} \otimes {\bf K}_{1_1}'\\
{\bf F}_{y_2} \otimes {\bf K}_{1_2}' \quad {\bf L}_{1_2} \otimes {\bf F}_{x_2}' \quad {\bf L}_{1_2} \otimes {\bf K}_{1_2}'\\
{\bf F}_{y_3} \otimes {\bf K}_{1_3}' \quad {\bf L}_{1_3} \otimes {\bf F}_{x_3}' \quad {\bf L}_{1_3} \otimes {\bf K}_{1_3}'\\
\end{bmatrix} 
\begin{bmatrix}
{\bf a} \\
{\bf b} \\
{\bf c}\\
\end{bmatrix} -
\begin{pmatrix}
\begin{bmatrix}
{\bf A}_{s_1}\\
{\bf A}_{s_2}\\
{\bf A}_{s_3}\\
\end{bmatrix}
 -
\begin{bmatrix}
{\bf F}_{y_1} \otimes {\bf F}_{x_1}'\\
{\bf F}_{y_2} \otimes {\bf F}_{x_2}'\\
{\bf F}_{y_3} \otimes {\bf F}_{x_3}'\\
\end{bmatrix} {\bf A}
\end{pmatrix}
}^2
\end{equation}

\noindent where $\hat{{\bf a}}, \hat{{\bf b}}, \hat{{\bf c}}$ are estimated high-pass subbands of the underlying unknown reference HR image. Once this system is solved, inverse Haar transform utilizing the reference LR image and the estimated highpass subbands gives the reconstructed HR image. 

The proposed algorithm can also be explained step by step as in {\bf Algorithm} - Super Resolution Reconstruction of Pansharpened MS images.

\begin{center}
	\begin{table}[h]
		{\bf Algorithm} \textit{SRR of Pansharpened MS images}
		\begin{itemize}
			\item \textit{Input}: Multitemporal MS and PAN bands
			\item \textit{Objective}: Reconstruct high resolution MS images, exceeding the PAN band spatial resolution
			\item \textit{Output}: High spatial/spectral resolution MS images
		\end{itemize}
		\begin{itemize} 
			\item[$\blacktriangleright$] Pansharpening
			\begin{itemize}
				\item[$\diamond$] Perform pansharpening using AWLP
			\end{itemize}  
			\item[$\blacktriangleright$] SRR Process --- do for each pansharpened MS band
			\begin{itemize}
				\item[$\diamond$] Image registration
				\begin{itemize}
				\item Choose a reference image
				\item Register all target images to reference
				\item Eliminate rotation for all target images
				\end{itemize} 
				\item[$\diamond$] Image reconstruction
				\begin{itemize}
				\item Construct coefficient matrices defined in Section \ref{sec:shift} for all LR images, using registration parameters found for translation
				\item Initialize HR image
				\item Do
				\begin{itemize}
					\item Perform Haar transform on HR image
					\item Use wavelet subbands with constructed matrices to estimate LR images
					\item Update HR using Eq. (10)
				\end{itemize}
				\item while $MSE({\bf A}_k - \hat{{\bf A}}_k) > \tau$, where $\tau$ is a predefined tolerance for error.
				\end{itemize} 
			\end{itemize}
		\end{itemize}
	\end{table}
\end{center}

\begin{figure}[h]
\centering
\begin{tabular}{ccc}
		\includegraphics[width=5cm]{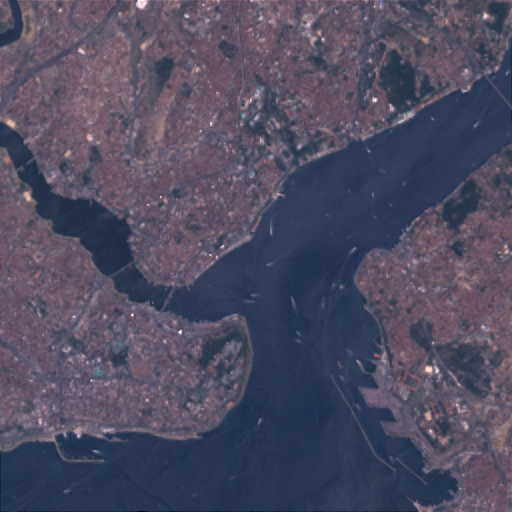} &
		\includegraphics[width=5cm]{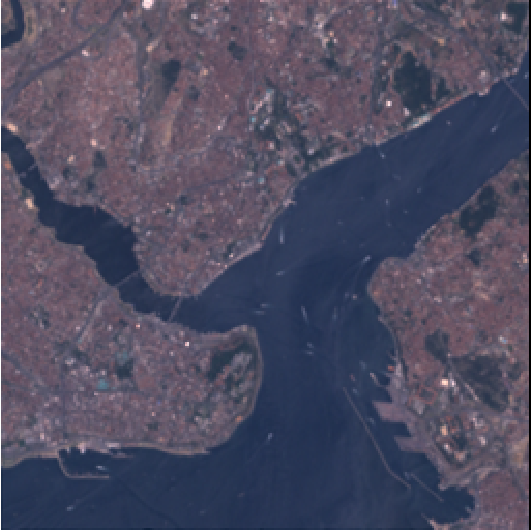} &
		\includegraphics[width=5cm]{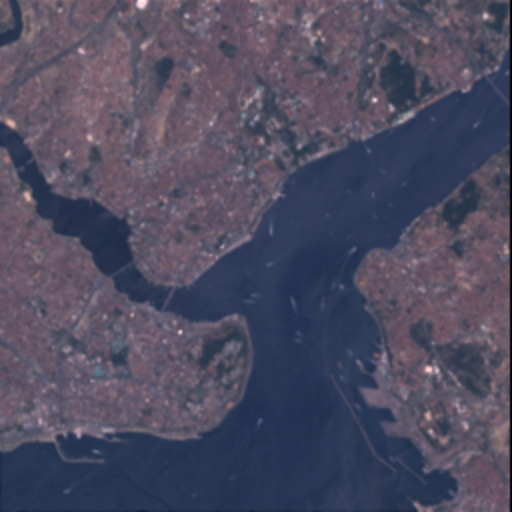} \\
		{({\bf a}) HR reference} & {({\bf b}) LR reference} & {({\bf c}) Linear} \\
		\\
		\includegraphics[width=5cm]{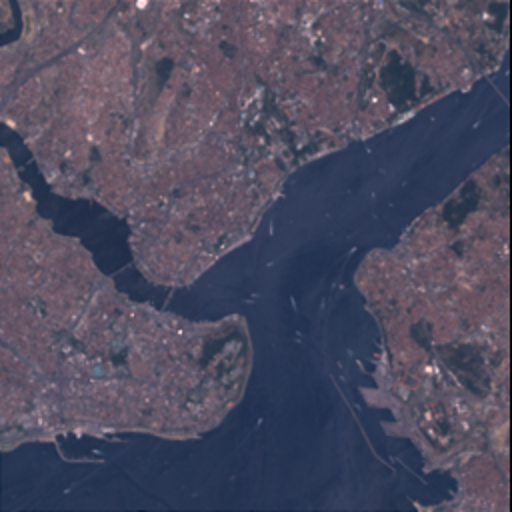} &
		\includegraphics[width=5cm]{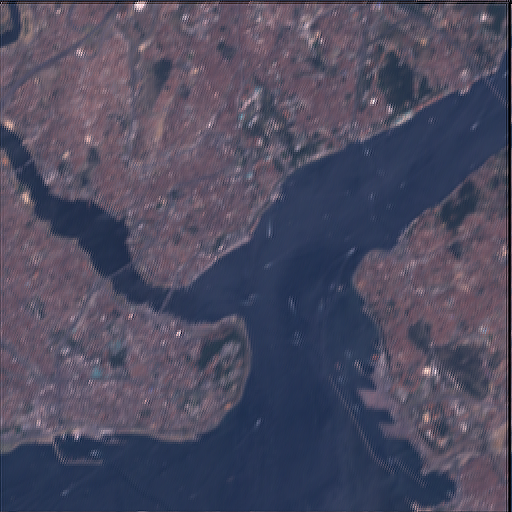} &
		\includegraphics[width=5cm]{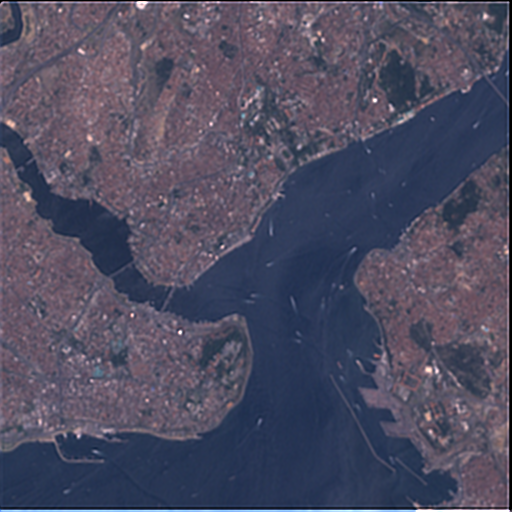} \\
		{({\bf d}) Bicubic} & {({\bf e}) IBP \cite{irani1991improving}} & {({\bf f}) Ours}
\end{tabular}
	\caption{Simulated results comparison of different methods with the reference image See the text and Table \ref{compPSNR} for more details.}
	\label{fig:simulated}
\end{figure} 

\begin{figure}[h]
\centering
\begin{tabular}{ccc}
		\includegraphics[width=5cm]{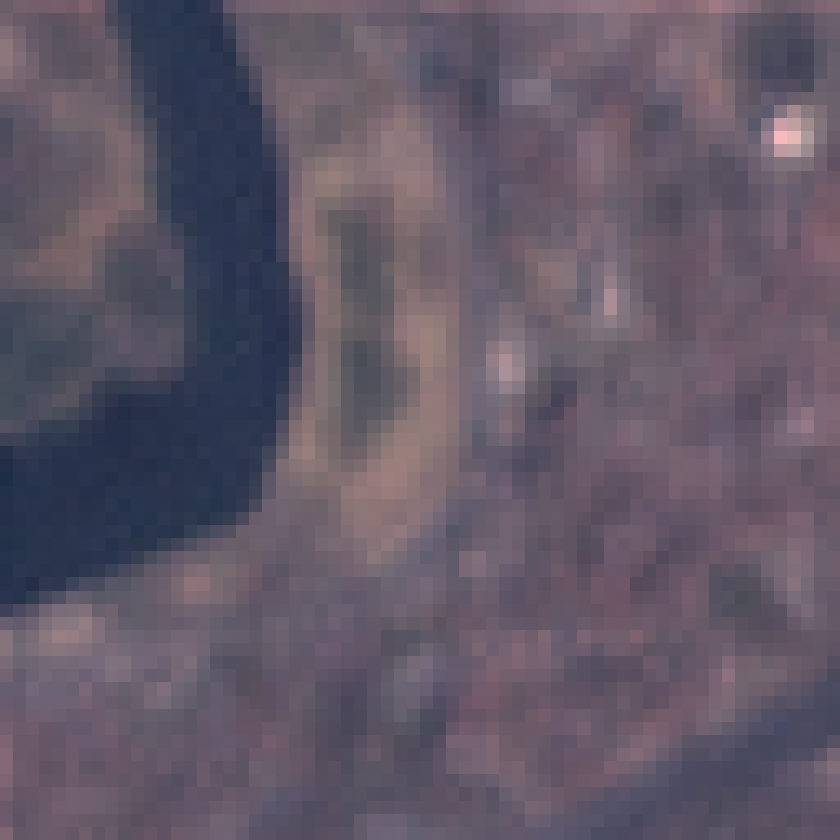} &
		\includegraphics[width=5cm]{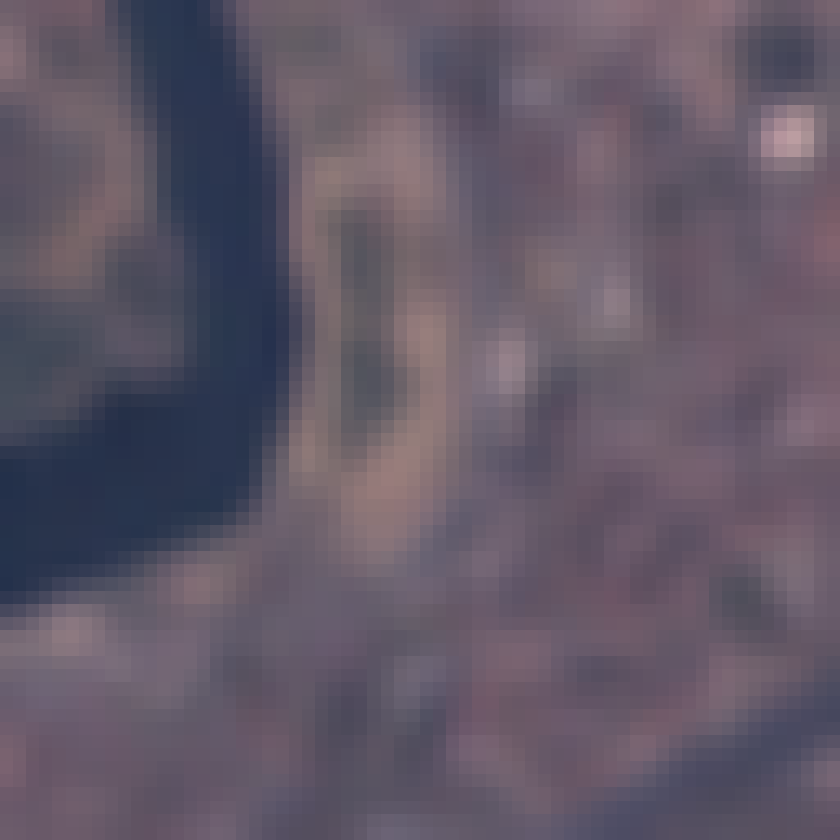} &
		\includegraphics[width=5cm]{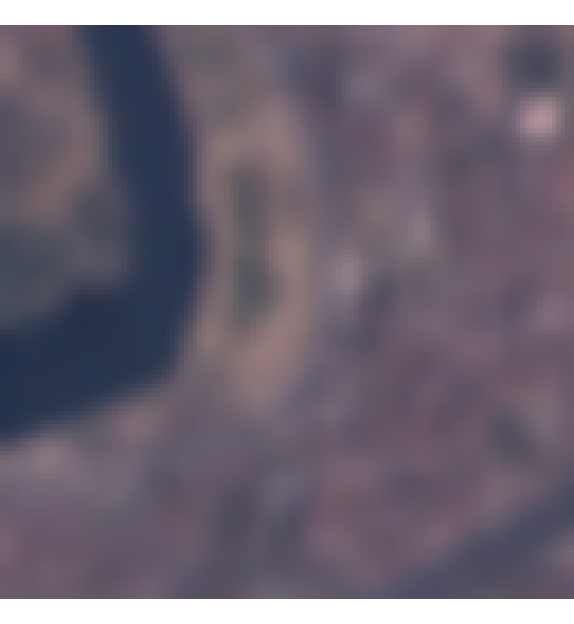} \\
		{({\bf a}) HR reference} & {({\bf b}) LR reference} & {({\bf c}) Linear} \\
		\\
		\includegraphics[width=5cm]{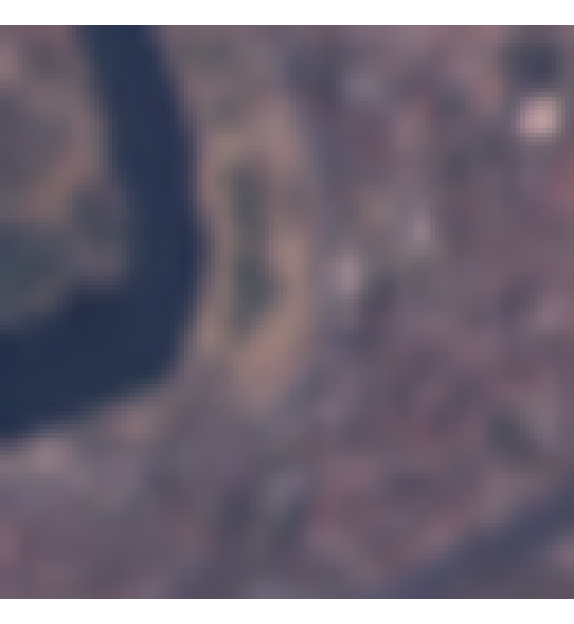} &
		\includegraphics[width=5cm]{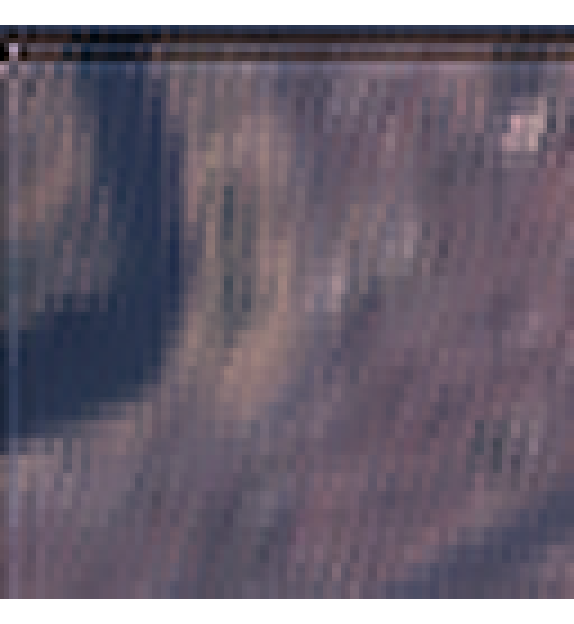} &
		\includegraphics[width=5cm]{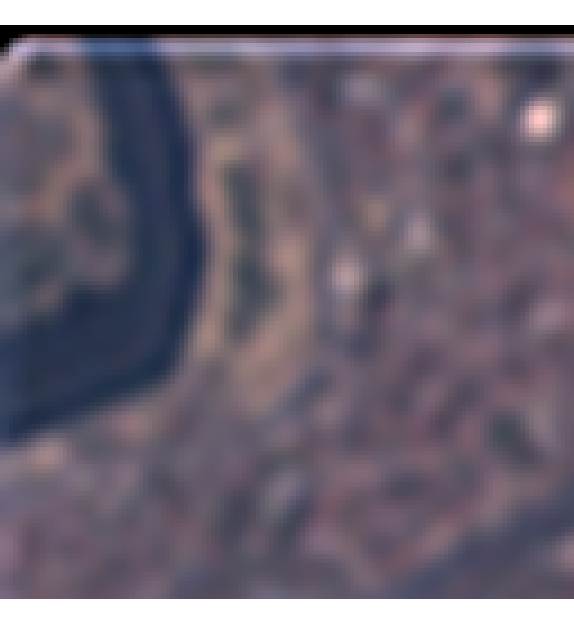} \\
		{({\bf d}) Bicubic} & {({\bf e}) IBP \cite{irani1991improving}} & {({\bf f}) Ours}
\end{tabular}
	\caption{Simulated results: comparison in a zoomed-in area. See the text and Table \ref{compPSNR} for more details.}
	\label{fig:simulated_zoom}
\end{figure}

\section{Experimental Results} \label{sec:exp}
We test our proposed method using Landsat 7 ETM+ images taken at different dates, which have seven MS bands together with a PAN band. The spectral resolution of MS bands range from $0.45 \mu$m to $2.35 \mu$m, while PAN bands span $0.52 - 0.9 \mu$m spectrum; and the spatial resolution of MS bands are $30$ m, whereas PAN bands are $15$ m. We select four multispectral image groups together with their PAN bands from a region in Marmara Sea, captured on July 2, September 4, in 2000, and May 18, August 6, in 2001. We conduct two sets of tests which are categorized as simulated and real experiments. All tests are carried after MS and PAN bands are fused in pansharpening. 

\subsection{Simulated Dataset}
Our first test is a simulated experiment, where one of the pansharpened MS images is chosen as reference, shifted in horizontal, vertical, and diagonal directions for one pixel, convolved with a Gaussian filter, and downsampled, which is a conventional method used for simulated SRR experiments \cite{zhang2012super}. The pansharped MS image chosen as reference is then used as the ground truth in comparisons.

For image registration, one of the bands of the reference pansharped MS image is chosen as the reference band, and image registration is performed for the same band of all datasets. Rotation is recovered for all target image bands, in order to have only translation between LR sets. Since we define our in-band shift method using subpixels and circular shifts, image regions are adjusted in order to cover the same area. We then initialize our HR estimate using the inverse transform of known reference LR and upsampled wavelet subbands of the LR image. The iterative method described in Section \ref{sec:sr} is then applied in order estimate the final HR image. We compare our results both qualitatively and quantitatively with conventional interpolation techniques and the IBP method \cite{irani1991improving}, since we propose a modified IBP model. All compared methods are given the same pansharpened MS images as input. Quantitative comparisons are based on Peak-Signal-to-Noise-Ratio (PSNR), Mean Square Errror (MSE) and Structural Similarity Index (SSIM).

Fig. \ref{fig:simulated} (a) shows the reference image used in simulated tests, which is from a region in Istanbul, Turkey, captured on July 2, 2000. All figures for simulated and real experiments, show a composite of R, G, and B bands.\\

Fig. \ref{fig:simulated} shows reference HR and reference LR images together with the compared methods including Bilinear interpolation, Bicubic interpolation and IBP method. In order to comprehend the results, \ref{fig:simulated_zoom} provides zoomed in areas of all images in \ref{fig:simulated}. As can be seen from these figures, the proposed method preserves spectral information of the MS bands while increasing the spatial resolution. Figures confirm that the proposed method reconstructs edges better than the compared ones.

Table \ref{compPSNR} provides the quantitative comparisons based on PSNR, MSE and SSIM values for each band, for a resolution enhancement factor of two. Since pansharpening methods use MS bands numbered $1,2,3,4,5,$ and $7$ for Landsat 7 ETM+, we compare the results for these bands. Quantitative comparisons in the table also validate the qualitative illustrations in the figures. Even though for some bands, the results obtained with the proposed method is lower than the compared ones, in general the proposed method preserves the spectral information better while increasing the spatial resolution.\\

\begin{center}
	\begin{table*}[t]
		\centering
		\caption{Comparison of proposed method with other methods in PSNR, MSE, SSIM.}
		\begin{tabular}{ r|rrr|rrr|rrr|rrr }
			\hline
			\multirow{2}{*}{Band} & \multicolumn{3}{c}{Linear} & \multicolumn{3}{c}{Bicubic} & \multicolumn{3}{c}{IBP} & \multicolumn{3}{c}{Proposed} \\
			\cline{2-13}
			& PSNR & MSE & SSIM & PSNR & MSE & SSIM & PSNR & MSE & SSIM & PSNR & MSE & SSIM \\
			\cline{1-13}
			1 & 33.57 & 0 & 0.88 & 33.94 & 0 & 0.91 & 25.71 & 0.004 & 0.58 & 34.82 & 0 & 0.91 \\
			2 & 27.24 & 0.004 & 0.78 & 27.37 & 0.004 & 0.81 & 25.17 & 0.006 & 0.58 & 25.70 & 0.004 & 0.84\\
			3 & 22.37 & 0.01 & 0.7 & 22.39 & 0.01 & 0.73 & 21.65 & 0.01 & 0.53 & 27.26 & 0.003 & 0.78\\
			4 & 20.18 & 0.02 & 0.49 & 20.19 & 0.02 & 0.5 & 19.92 & 0.02 & 0.42 & 16.47 & 0.03 & 0.51\\
			5 & 25.18 & 0.006 & 0.73 & 25.09 & 0.006 & 0.74 & 22.99 & 0.01 & 0.45 & 26.01 & 0.005 & 0.7\\
			7 & 25.21 & 0.006 & 0.72 & 25.13 & 0.006 & 0.73 & 23.01 & 0.01 & 0.44 & 26.03 & 0.005 & 0.72\\
			\bottomrule
		\end{tabular} \label{compPSNR}
	\end{table*}
\end{center}

\subsection{Real Dataset}
\begin{figure}[h]
\begin{center}
\begin{tabular}{cc}
		\includegraphics[width=0.4\linewidth]{reference.png} &
		\includegraphics[width=0.4\linewidth]{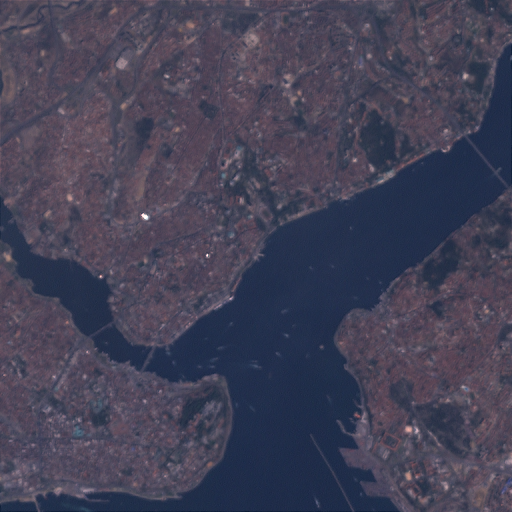}\\
		\includegraphics[width=0.4\linewidth]{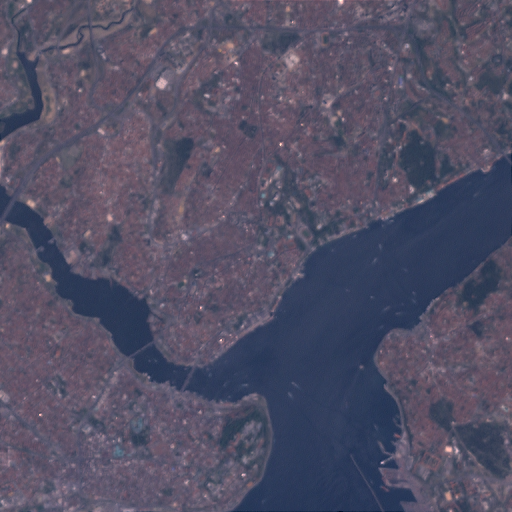}&
		\includegraphics[width=0.4\linewidth]{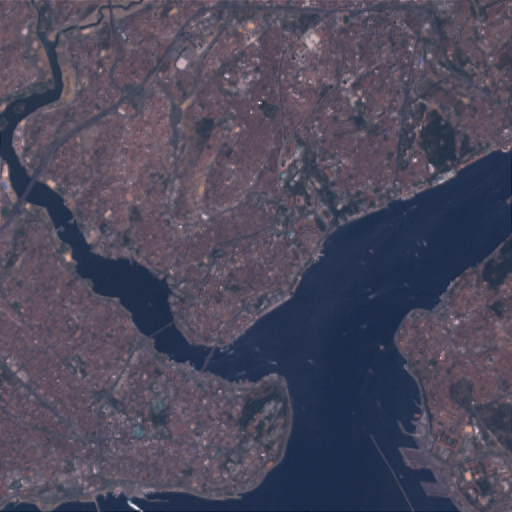}
\end{tabular}
		\caption{Real pansharped MS images (RGB bands).} 
		\end{center} \label{fig:real}
\end{figure}

For the second test, we use pansharpened MS bands as our LR image set, and estimate an HR image without a ground truth. This test is compared qualitatively only. Fig. \ref{fig:real} shows the real dataset used for the experiments.

\section{Conclusion} \label{sec:conc}
Multispectral sensors provide low-spatial MS and high-spatial PAN images. Pansharpening or SRR methods are used separately in order to obtain high-spatial resolution MS bands without losing spectral information. In this paper, we propose employing pansharpening and SRR method together, to exceed the spatial resolution available in the PAN bands, by using both spatial and temporal information captured by most multispectral sensors. We perform pansharpening and registration before applying the proposed SRR technique, for which we derive closed-form model of wavelet domain in-band relationships. Experimental results demonstrate that the proposed scheme indeed exceeds the spatial resolution of PAN bands, while keeping the spectral information of MS bands.


{
\bibliographystyle{plain}
\bibliography{foroosh,egbib}
}

\end{document}